\documentclass[sigconf]{acmart}

\usepackage{amsmath}
\usepackage{algorithm}
\usepackage{algorithmic}
\usepackage{graphicx}
\usepackage{subfigure}
\usepackage{multirow}
\usepackage{diagbox}
\usepackage{makecell}

\AtBeginDocument{%
  \providecommand\BibTeX{{%
    \normalfont B\kern-0.5em{\scshape i\kern-0.25em b}\kern-0.8em\TeX}}}

\copyrightyear{2023} 
\acmYear{2023} 
\setcopyright{acmlicensed}\acmConference[WWW '23]{Proceedings of the ACM Web Conference 2023}{May 1--5, 2023}{Austin, TX, USA}
\acmBooktitle{Proceedings of the ACM Web Conference 2023 (WWW '23), May 1--5, 2023, Austin, TX, USA}
\acmPrice{15.00}
\acmDOI{10.1145/3543507.3583499}
\acmISBN{978-1-4503-9416-1/23/04}

\begin{document}

\title{Generating Counterfactual Hard Negative Samples for Graph Contrastive Learning}

\author{Haoran Yang}
\authornote{Both authors contributed equally to this research.}
\email{haoran.yang-2@student.uts.edu.au}
\affiliation{%
	\institution{University of Technology Sydney}
	\city{Sydney}
	\state{NSW}
	\country{Australia}
}

\author{Hongxu Chen}
\authornotemark[1]
\authornote{Corresponding author.}
\email{hongxu.chen@uts.edu.au}
\affiliation{%
  \institution{University of Technology Sydney}
  \city{Sydney}
  \state{NSW}
  \country{Australia}
}

\author{Sixiao Zhang}
\email{zsx57575@gmail.com}
\affiliation{%
\institution{University of Technology Sydney}
	\city{Sydney}
  \state{NSW}
  \country{Australia}
}

\author{Xiangguo Sun}
\email{xiangguosun@cuhk.edu.hk}
\affiliation{%
	\institution{The Chinese University of Hong Kong}
	\city{Hong Kong SAR}
	\country{China}
}

\author{Qian Li}
\email{qli@curtin.edu.au}
\affiliation{%
	\institution{Curtin University}
	\city{Perth}
	\state{WA}
	\country{Australia}
}

\author{Xiangyu Zhao}
\email{xianzhao@cityu.edu.hk}
\affiliation{
	\institution{City University of Hong Kong}
	\city{Hong Kong SAR}
	\country{China}
}

\author{Guandong Xu}
\authornotemark[2]
\email{guandong.xu@uts.edu.au}
\affiliation{%
	\institution{University of Technology Sydney}
	\city{Sydney}
	\state{NSW}
	\country{Australia}
}

\renewcommand{\shortauthors}{Yang and Chen, et al.}
\newtheorem{myDef}{Definition}

\begin{abstract}
Graph contrastive learning has emerged as a powerful unsupervised graph representation learning tool. The key to the success of graph contrastive learning is to acquire high-quality positive and negative samples as contrasting pairs to learn the underlying structural semantics of the input graph. Recent works usually sample negative samples from the same training batch with the positive samples or from an external irrelevant graph.
However, a significant limitation lies in such strategies: the unavoidable problem of sampling false negative samples. In this paper, we propose a novel method to utilize \textbf{C}ounterfactual mechanism to generate artificial hard negative samples for \textbf{G}raph \textbf{C}ontrastive learning, namely \textbf{CGC}. We utilize a counterfactual mechanism to produce hard negative samples, ensuring that the generated samples are similar but have labels that differ from the positive sample. The proposed method achieves satisfying results on several datasets. It outperforms some traditional unsupervised graph learning methods and some SOTA graph contrastive learning methods. We also conducted some supplementary experiments to illustrate the proposed method, including the performances of CGC with different hard negative samples and evaluations for hard negative samples generated with different similarity measurements.
The implementation code is available online to ease reproducibility\footnote{https://www.dropbox.com/sh/kyf8p9unkhn0r99/AABd33jFBfjGYIkvIqWpuNwYa?dl=0}.
\end{abstract}



\maketitle

\section{Introduction}
Graph contrastive learning (GCL) \cite{dgi, gcc, graph-aug, graph-ada-aug, mvgrl, dsgc, infograph} has emerged as a powerful learning paradigm for unsupervised graph representation learning. Inspired by the widely adopted contrastive learning framework in computer vision (CV) \cite{cl-cv, moco} and natural language processing (NLP) \cite{electra}, GCL leverages the advanced representation learning capabilities of graph neural networks (GNNs) \cite{gcn, gin, jin2022automated, fan2022graph} and tries to distil high-quality representative graph embeddings of an input graph via comparing the differences and similarities among augmented graphs (i.e., positive and negative samples) derived from the original input.

The key to a successful GCL method is to derive high-quality contrasting samples from the original input graph. To date, various kinds of methods to generate positive samples are proposed, for example, graph augmentations-based approaches \cite{graph-aug, graph-ada-aug} and multi-view sample generation \cite{mvgrl, dsgc}, which have been becoming dominant and achieved satisfying performance. Despite this progress, especially in manipulating positive pairs, far less attention has been given to obtaining negative samples \cite{hard-negative}. Compared to positive samples in contrastive learning, negative sampling is more challenging and non-trivial\cite{hard-negative}. Existing methods of negative sample acquisition mainly follow traditional sampling techniques, which may encounter the deficiency caused by unnoticeable false negative samples \cite{dgcl}. For instance, GraphCL \cite{graph-aug} samples other graphs as the negative samples from the same training batch where the target graph comes from. Such an approach does not guarantee that the sampled negative graphs are true. GCC \cite{gcc} samples negative graphs from an external graph based on the assumption that common graph structural patterns are universal and transferable across different networks. However, this assumption neither has any theoretical guarantee nor has been validated by empirical study\cite{gcc}. To alleviate the impact caused by false-negative samples, debiasing treatment has been introduced to current graph contrastive learning methods \cite{dgcl, gdcl}. The idea of these debiased graph contrastive learning methods is to estimate the probability of whether a negative sample is false. Based on this, some negative samples with low confidence will be discarded or treated with lower weights in the contrastive learning phase. Nevertheless, a typical major limitation of both GCL and the debiasing-based variants is still evident - most of these sampling strategies are stochastic and random. In other words, current methods do not guarantee the quality of the sampled negative pairs.

To address the previously mentioned problems, we first name the high-quality negative samples as hard negative samples and give corresponding definitions. According to \cite{hard-negative}, a hard negative sample is a data instance whose label differs from that of the target data, and its embedding is similar to that of the target data. Considering the limitations of sampling-based strategies discussed previously, we argue that a strictly constrained generative process must be imposed to guarantee the quality of the negative samples (a.k.a. generating hard negative samples). 
Inspired by counterfactual reasoning \cite{counterfactual-survey}, a fundamental reasoning pattern of human beings, which helps people to reason out what minor behaviour changes may result in considerable differences in the final event outcome. We intuitively came up with the idea that the hard negative sample generation should apply minor changes to the target graph and finally can obtain a perturbed graph whose label is strictly different from the original graph.


\begin{figure}[htbp]
	\centering
	\includegraphics[width=0.45\textwidth]{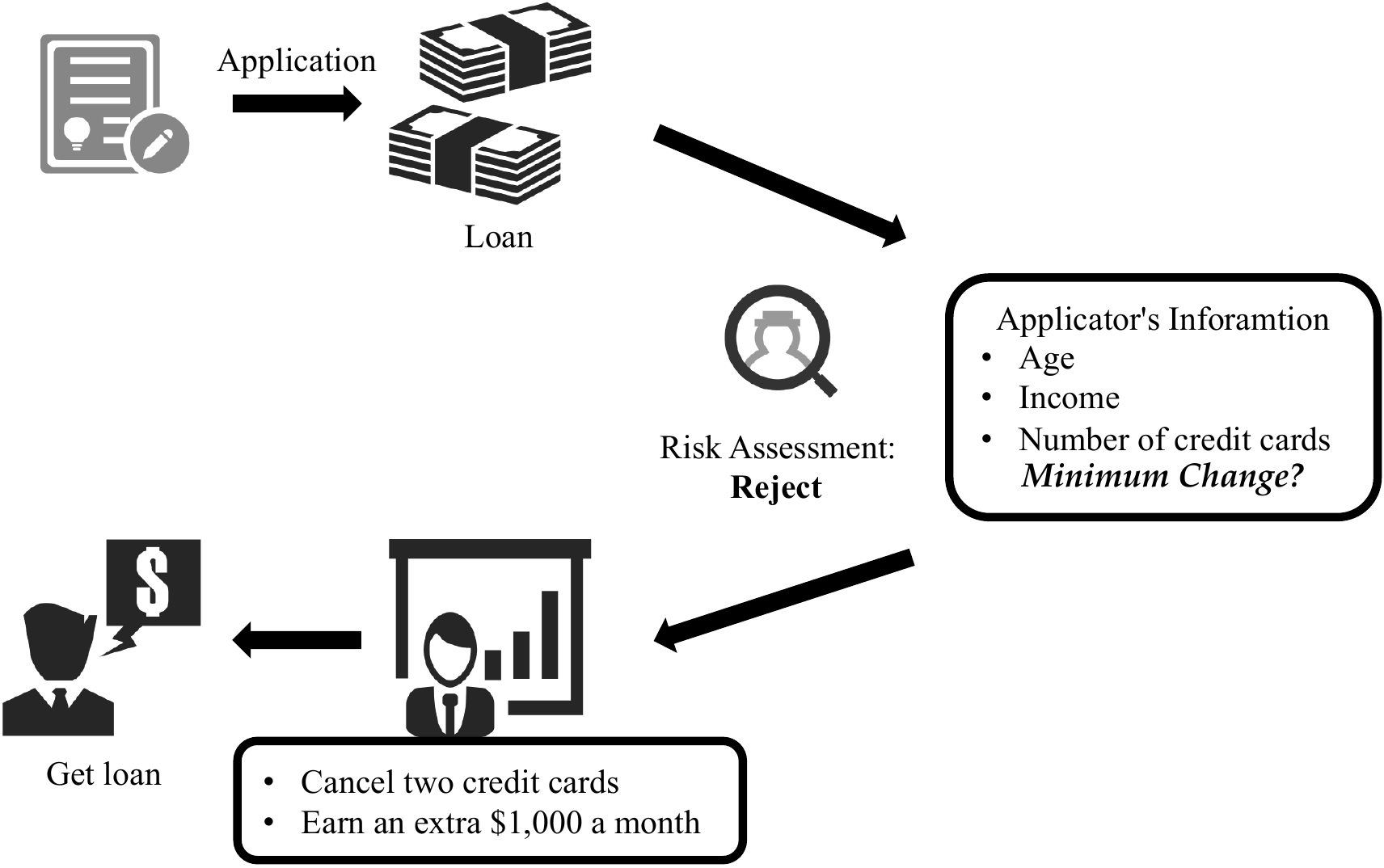}
	\caption{An illustrative example about a counterfactual explanation. Someone wants to apply for a loan, but the application is rejected after a risk assessment from the financial institution. Many factors are related to the final decision, such as the applicant's age, income, and number of credit cards. The minimum change the applicant needs to make to get the loan is earning an extra \$1,000 per month or cancelling two credit cards.}
        \Description{An illustrative example about a counterfactual explanation. Someone wants to apply for a loan, but the application is rejected after a risk assessment from the financial institution. Many factors are related to the final decision, such as the applicant's age, income, and number of credit cards. The minimum change the applicant needs to make to get the loan is earning an extra \$1,000 per month or cancelling two credit cards.}
	\label{fig:counterfactual}
 \vspace{-3mm}
\end{figure}

To this end, we propose generating two types of hard negative samples via perturbations to the original input graph: proximity perturbed graphs and feature-masked graphs.  It is worth noting that these two types of generation processes will be adaptively conducted and constrained by sophisticating similarity-aware loss functions. 
However, this process is still challenging and non-trivial. We believe there are two significant challenges. First, in graph perturbation and feature masking, how to measure a generated sample is \textit{hard}? 
To solve the problem, we first design two indication matrices demonstrating the changes made to the graph structure and feature space. Then, different matrix norms are applied to indication matrices to reflect how much perturbation has been made. We will minimise the calculated matrix norms such that the perturbation applied to the original graph to generate negative samples is as minor as possible. In this case, the generated samples would be similar to the original graph in proximity and feature space. By adopting matrix norms, we can quantify the perturbation and ensure that the generated samples are \textit{hard} ones.
After formulating a constraint that forces the generated samples to be hard to distinguish from the target in proximity and feature space, the second challenge is how to make sure the generated hard samples have different labels from the target. That is to say, how can we ensure the generated hard samples are \textit{true negative}? 
We first feed the target graph and the generated samples into a graph classifier to achieve that. The classifier will then output the probability distributions of the classes to which the target graph and the generated samples belong. Following the counterfactual mechanism, an objective measuring the differences between the classifier's outputs for the target graph and that for the generated samples is applied and minimised. Specifically, the similarities between the predicted probability distributions are minimised via monitoring the KL divergence.
With the two objectives described above, we can generate high-quality hard negative samples with proper and reasonable constraints.


We propose a counterfactual-inspired generative method for graph contrastive learning to obtain hard negative samples. It explicitly introduces constraints to ensure the generated negative samples are true and hard, eliminating the random factors in current negative sample acquiring methods in GCL methods. Furthermore, once the generation procedure is finished, we do not need further steps (e.g., debiasing) to process the acquired samples. The contributions of our work are summarized as follows:
\begin{itemize}
	\item We propose a novel adaptively graph perturbation method, CGC, to produce high-quality hard negative samples for the GCL process.
	\item We innovatively introduce the counterfactual mechanism into the GCL domain, leveraging its advantages to make the generated negative samples be \textit{hard} and \textit{true}. Due to the successful application of the counterfactual mechanism in our work, there is potential feasibility of conducting the counterfactual reasoning to explain GCL models.
	\item We conducted extensive experiments to demonstrate the proposed method's effectiveness and properties, which achieved state-of-the-art performances compared to several classic graph embedding methods and some novel GCL methods.
\end{itemize}

\begin{figure*}[htbp]
	\centering
	\includegraphics[width=0.7\textwidth]{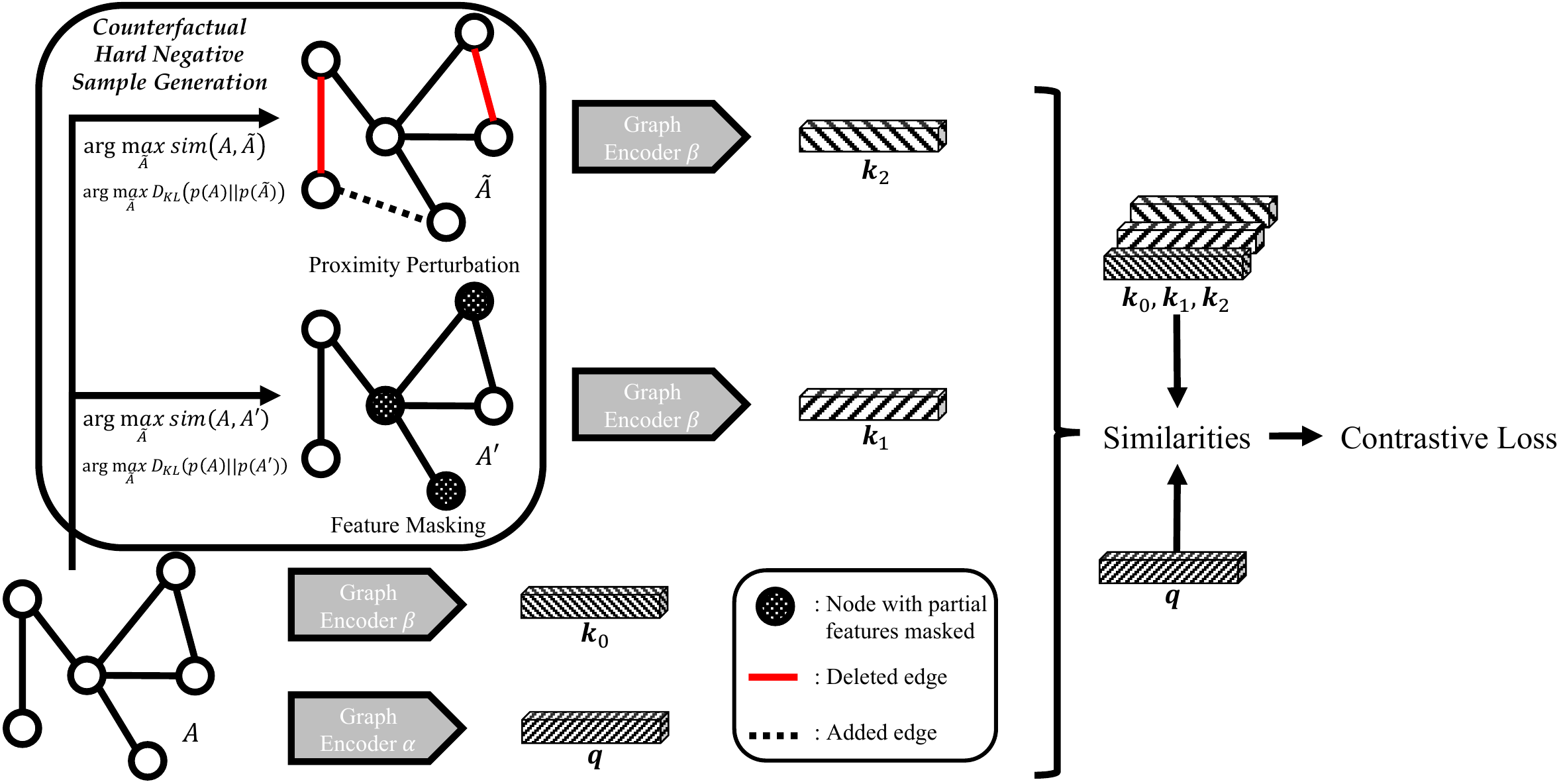}
	\caption{The overview of CGC. We first conduct counterfactual hard negative sample generation to acquire a proximity-perturbed and feature-masked sample. Then, the target and the two generated hard negative samples will be fed into the graph contrastive learning module to learn graph embeddings.}
        \Description{The overview of CGC. We first conduct counterfactual hard negative sample generation to acquire a proximity-perturbed and feature-masked sample. Then, the target and the two generated hard negative samples will be fed into the graph contrastive learning module to learn graph embeddings.}
	\label{fig:overview}
 \vspace{-3mm}
\end{figure*} 

\section{Related Work}
This section briefly introduces the backgrounds and research progress in two highly related areas, including counterfactual reasoning mechanisms and graph contrastive learning.
\subsection{Counterfactual Reasoning Mechanism}
\label{sec:counterfactual}
Counterfactual reasoning is a basic way of reasoning that helps people understand their behaviours and the world's rules \cite{counterfactual-survey}. The definition of counterfactual reasoning is given by \cite{counterfactual-def} stating that counterfactual is a probabilistic answer to a `what would have happened if' question. Many illustrative examples are provided in \cite{counterfactual-tutorial} to help understand the ideas behind counterfactual. For instance, as shown in Figure \ref{fig:counterfactual}. Counterfactual is a kind of thinking mechanism to discover the facts that contradict existing facts and could potentially alter the outcomes of a decision-making process. There are some restrictions on counterfactuals. First, many factors could potentially affect the final results. However, counterfactuals must apply as small as possible changes to achieve such a goal. Second, counterfactuals must be feasible and reasonable. In Figure \ref{fig:counterfactual}, the financial institution would release the loan without hesitation if the applicator earns an extra one million dollars per month. Nevertheless, the applicator cannot have such a high salary quickly. So, earning an extra one million dollars per month is not a counterfactual. A classical counterfactual method is heuristic counterfactual generation \cite{counterfactual-alg}, shown in Algorithm \ref{alg:counterfactual}, where
\begin{equation}
	L(x, x', y', \lambda)=\lambda\overbrace{(\hat{f}(x')-y')^2}^{\text{distance in predictions}}+\underbrace{d(x, x')}_{\text{distance in instances}},
\end{equation}
and $x$ denotes the target instance, $x'$ is counterfactual, $y'$ represents the desired outcome, $\lambda$ is the term used to balance two distances, and $\epsilon$ denotes tolerance for the distance. This equation is the objective function of the heuristic counterfactual generation algorithm. It maximises the distances in predictions and minimises the distance between the original instance $x$ and the counterfactual $x'$.

\begin{algorithm}
	\caption{Heuristic counterfactual generation algorithm}
	\label{alg:counterfactual}
	\begin{algorithmic}
		\STATE{sample a random instance as the initial $x'$}
		\STATE{optimise $L(x, x', y', \lambda)$ with $x'$}
		\WHILE{$|\hat{f}(x')-y'|>\epsilon$}
		\STATE{increase $\lambda$ by step-size $\alpha$}
		\STATE{optimise $L(x, x', y', \lambda)$ with new $x'$}
		\ENDWHILE
		\RETURN{$x'$}
	\end{algorithmic}
\end{algorithm}

\subsection{Graph Contrastive Learning}

Recently, many researchers devoted themselves to constructing proper positive pairs to conduct graph contrastive learning. 
There are plenty of works describing how to generate high-quality positive pairs to conduct graph contrastive learning \cite{dgi, infograph, graph-aug, mvgrl, dsgc, gcc}, and indeed, they have achieved promising performances. However, works introducing how to obtain negative samples to facilitate graph contrastive learning further are scarce in the current literature. Recent works usually adopted sampling strategies to acquire negative samples. Specifically, in GraphCL \cite{graph-aug}, the authors proposed to sample other graphs as the negative samples from the same training batch in which the target graph is. However, under the scenario of lacking label information, the sampled graphs may have the same label as the target graph, resulting in sampling false negative graphs. Similarly, GCA and InfoGraph conducted contrastive learning at the node level. They sampled negative nodes from the neighbours of the target \cite{graph-ada-aug, infograph}. Such a strategy would also meet the false negative sample problem. Moreover, GCC was proposed to sample negative graphs from an external dataset. Undoubtedly, the external negative samples have different semantics or labels from the target graph \cite{gcc}. However, such a strategy was proposed based on the hypothesis that representative graph structural patterns are universal and transferable across networks \cite{gcc}. This hypothesis has not been verified that it holds among all graph datasets. 

Sampling-based methods for hard negative sample generation suffer from the false negative sample problem and attract the researchers' attention. Some recent works \cite{dgcl, gdcl} tried to address such a problem by relieving the biases existing in the negative graph sampling process. DGCL \cite{dgcl} found out the \textit{harder} the negative sample is, the more likely a false negative sample is, and the probability of being a false negative sample is related to the similarity between the target and the sampled negative instances. The strategy DGCL adopted is straightforward to understand, reducing the weight of the negative samples that are likely to be false in the contrastive training procedure. Such a method indeed relieved the impact brought by the false negative samples. GDCL \cite{gdcl} utilised graph clustering techniques to determine if a negative sample is false before feeding it into the contrastive learning process. Graph clustering is applied to the set containing the initial sampled negative instances and the target. The instances in the cluster where the target is would be discarded because they are more likely to have the same labels as the target. 

Nevertheless, a typical major limitation of both GCL and the debiasing-based variants is still evident - most of these sampling strategies are stochastic and random. In other words, current methods do not guarantee the quality of the sampled negative pairs.

\section{Methodology}
This section will give a detailed illustration of the proposed method and the training procedure. The overview of the proposed method is illustrated in Figure \ref{fig:overview}.

\subsection{Problem Definition}
\label{sec:problem}
\begin{figure}[htbp]
	\centering
	\includegraphics[width=0.45\textwidth]{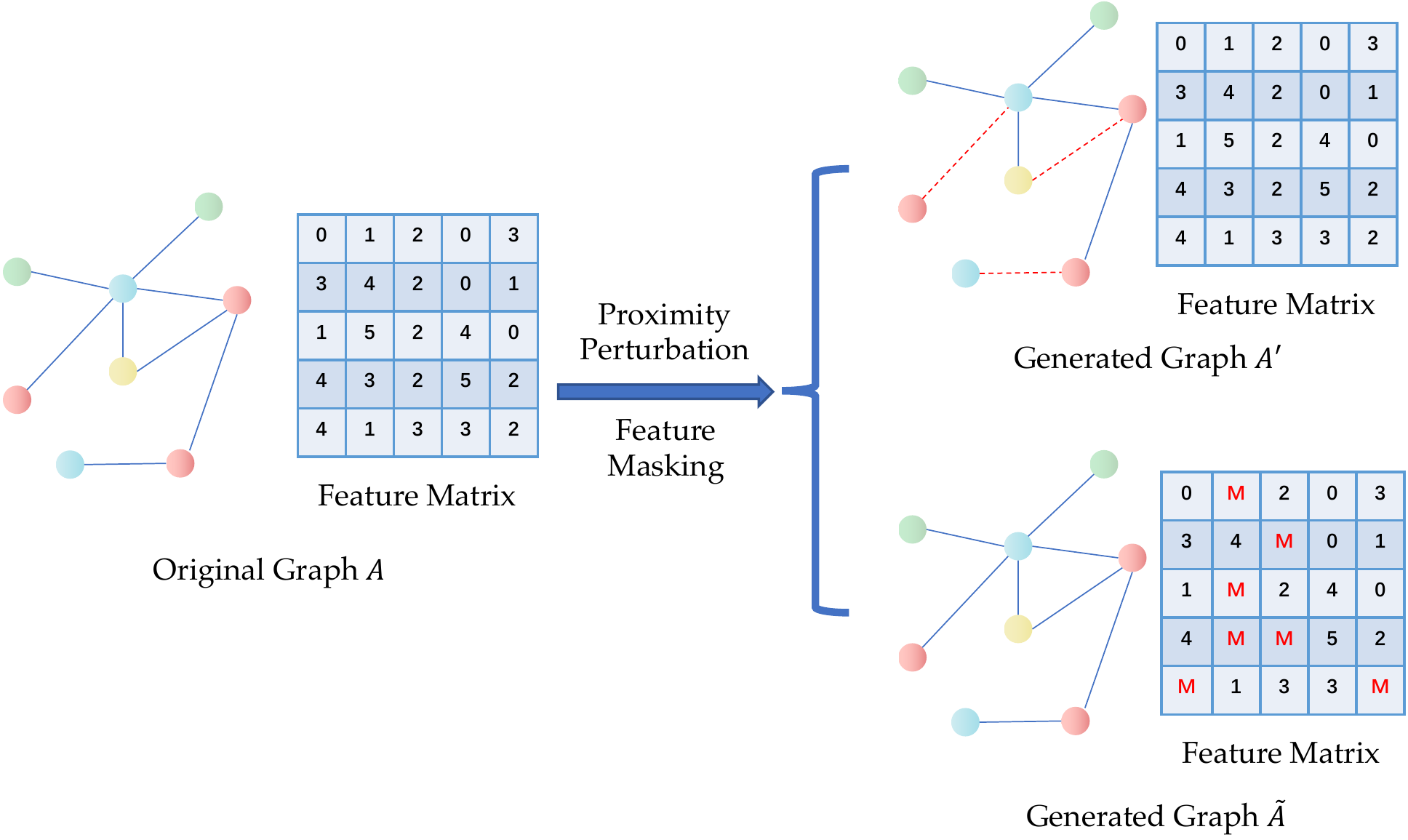}
	\caption{An example of the graphs with proximity perturbed and with features masked derived from an original graph.}
        \Description{A toy example about the graph with proximity perturbed and the graph with features masked derived from the original graph.}
	\label{fig:generation}
 \vspace{-3mm}
\end{figure}

Given a graph $\mathcal{A}=\{\mathcal{V}, \mathcal{E}, \mathcal{X}\}$, where $\mathcal{V}$ denotes all the nodes, $\mathcal{E}$ represents all the edges, and $\mathcal{X}$ is the set consists of the features of all nodes. If there are $N$ nodes and the dimension of the feature is $h$, then, $\mathcal{X}\in\mathbb{R}^{N\times h}$. Our method aims to derive some negative graphs from the input graph based on counterfactual mechanisms, and a toy example is shown in Figure \ref{fig:generation}. For simplicity's sake, in this paper, we consider the scenario where two kinds of  hard negative graphs are generated, the proximity perturbed graph $\mathcal{A}'=\{\mathcal{V}, \mathcal{E}', \mathcal{X}\}$ and the feature masked graph $\tilde{\mathcal{A}}=\{\mathcal{V}, \mathcal{E}, \tilde{\mathcal{X}}\}$, such that
\begin{equation}
	\mathop{\arg\max}_{\mathcal{E}', \tilde{\mathcal{X}}} sim(\mathcal{A}, \mathcal{A}')+sim(\mathcal{A}, \tilde{\mathcal{A}}),
\end{equation}
\begin{equation}
	\mathop{\arg\max}_{\mathcal{E}', \tilde{\mathcal{X}}} D_{KL}(p(\mathcal{A})||p(\mathcal{A}'))+D_{KL}(p(\mathcal{A})||p(\tilde{\mathcal{A}})),
\end{equation}
where $sim(*)$ denotes the metric to measure the similarity between two items (e.g., graph adjacency matrices, feature matrices), $D_{KL}(*)$ is the KL-Divergence \cite{kl} function, which is used to measure the similarity between two probability distributions, and $p(*)$ denotes predictor outputting the probabilities of classes to which the graph belongs. The intuition behind the two formulas is to derive hard negative graphs with different labels while forcing the derived graphs to be as similar to the original graph as possible. In other words, we want to achieve dramatic change at the semantics level with minor perturbations at the graph's essential elements. 
The problem is formulated as an optimisation problem to maximise the similarities between the generated negative samples and the target in proximity and feature and force them to have different labels.

\subsection{Counterfactual Adaptive Perturbation}
\begin{table*}[htbp]
\caption{Statistics of four graph datasets.}
	\label{tab:statistics}
  \vspace{-3mm}
	\begin{tabular}{c|ccccc}
		\toprule
		Dataset        & Num. of Graphs & Avg. Num. of Nodes & Avg. Num. of Edges & Node Attr. Dim. & Num. of Classes \\ \midrule
		PROTEINS\_full & 1,113          & 39.06              & 72.82              & 29              & 2               \\
		FRANKENSTEIN   & 4,337          & 16.90              & 17.88              & 780             & 2               \\
		Synthie        & 400            & 95.00              & 172.93             & 15              & 4               \\
		ENZYMES        & 600            & 32.63              & 62.14              & 18              & 6               \\ \bottomrule
	\end{tabular}
 \vspace{-3mm}
\end{table*}
This paper discusses two adaptive perturbation matrices: the proximity perturbation matrix and the feature masking matrix.

\subsubsection{\textbf{Proximity Perturbation}}
Proximity perturbation aims to change the graph structure to generate a contrasting sample. This can help the model learn the critical structural information in the original graph \cite{graph-aug, graph-ada-aug}. First, let us focus on how to conduct adaptive perturbation. To achieve the goal of adaptive perturbation, we need a trainable matrix $\mathbf{M}_a\in\mathbb{R}^{N\times N}$, such that
\begin{equation}
	\mathbf{A}_a=\mathbf{M}_a\times\mathbf{A},
\end{equation}
where $\mathbf{A}\in\mathbb{R}^{N\times N} $ is the adjacency matrix of $\mathcal{A}$, and $\mathbf{A}_a$ denotes the adjacency matrix of the proximity perturbed graph $\mathcal{A}'$. Note that we adopt matrix multiplication here instead of taking the Hadamard product since it cannot add an edge to the adjacency matrix. Moreover, values of the entries of  $\mathbf{A}_a$ are in $\mathbb{R}^{N\times N}$, which conflict with the definition domain of the adjacency matrix, $\{0, 1\}^{N\times N}$. We need an extra step such that $f:\mathbf{A}_a\in\mathbb{R}^{N\times N}\rightarrow\mathbf{A}^{'}_{a}\in\{0, 1\}^{N\times N}$. We use the following formula to conduct the mapping:
\begin{equation}
	\mathbf{A}^{'}_{a}=\mathbb{I}(sigmoid(\mathbf{A}_{a})\geq \omega),
\end{equation}
where $\mathbb{I}(*)$ is the indicator function, and $\omega$ is a threshold determining whether to set the entry as 1 or 0. Finally, we have the adjacency matrix for the negative sample $\mathcal{A}'$. Consequently, a perturbed set $\mathcal{E}'$ of edges is obtained. Though in this procedure, there is no modification made to nodes, we can adaptively discard some nodes by deleting all the edges of the nodes, which will be isolated from the generated graph after perturbation. In summary, we can utilise the proximity perturbation matrix and procedure mentioned above to simulate all the proximity perturbation methods in \cite{graph-aug}, including edge dropping or adding and node dropping. 

\subsubsection{\textbf{Feature Masking}}
Feature masking tries to mask the original feature matrix to help the contrastive learning model obtain the critical information in the features \cite{graph-aug, graph-ada-aug}, which would determine the decisive factors in the feature domain. The whole procedure of feature masking is similar to the proximity perturbation, but there are some minor differences. First, same as the previous procedure, we need to initialize a trainable matrix $\mathbf{M}_b\in\mathbb{R}^{N\times h}$, which serves as a mask matrix here. We have to make sure the definition domain of it is $\{0, 1\}^{N\times h}$. To achieve such a goal, we have
\begin{equation}
	\mathbf{M}^{'}_b=\mathbb{I}(sigmoid(\mathbf{M}_b)\geq \gamma),
\end{equation}
where $\gamma$ is a threshold determining whether to mask some feature entries. To fulfil the feature masking procedure, we need to have Hadamard element-wise product between $\mathbf{M}^{'}_b$ and $\mathcal{X}$ instead of matrix multiplication in the previous perturbation procedure:
\begin{equation}
	\tilde{\mathcal{X}} = \mathbf{M}^{'}_b\circ\mathcal{X}.
\end{equation}
Once we finish the feature masking procedure, the values of the masked features will be replaced with $0$.

After the proximity perturbation and the feature masking, two hard negative samples are acquired, which are $\mathcal{A}'=\{\mathcal{V}, \mathcal{E}', \mathcal{X}\}$ and $\tilde{\mathcal{A}}=\{\mathcal{V}, \mathcal{E}, \tilde{\mathcal{X}}\}$, respectively.

\subsubsection{\textbf{Perturbation Measurement}}
Once we obtained two perturbed graphs, the issue we have to deal with next is how to ensure the generated graphs are hard negatives. As we mentioned previously, we leverage the counterfactual mechanism to solve this problem because this method naturally meets the requirements of hard negative sample generation. Both aim to output something different at the semantic level but similar at the structural level. We introduced two objectives in the Section. \ref{sec:problem}, and more details are given in this section.

First, we discuss maximising the similarity between the original and the perturbed graphs. This objective function tries to ensure the perturbation we made is as minor as possible. We utilise the Frobenius norm of the difference between $\mathbf{A}$ and $\mathbf{A}'_a$. The smaller the Frobenius norm is, the more similar they are. The Frobenius norm of the masking matrix can measure the similarity between features. A relatively greater norm indicates that a small portion of feature entries are masked. Therefore, the similarity between $\mathcal{X}$ and $\tilde{\mathcal{X}}$ would be high. The objective is formulated as follows:
\begin{equation}
	\mathcal{L}_s = ||\mathbf{A}-\mathbf{A}^{'}_a||_F-||\mathbf{M}'_b||_F.
\end{equation}

Next, we must ensure the generated graphs are different from the original graph at the semantic level. Here, we consider the classification problem, where we minimize the similarity between probability distributions of classes between the original and the perturbed graphs. Therefore, we have
\begin{equation}
	\mathcal{L}_c=-D_{KL}(p(\mathcal{A}), p(\mathcal{A}'))-D_{KL}(p(\mathcal{A}), p(\tilde{\mathcal{A}})).
	\label{eq:hard}
\end{equation}  

So, the overall objective for counterfactual pre-training for hard negative sample generation is
\begin{equation}
	\mathcal{L}_{pre}=\mathcal{L}_s+\mathcal{L}_c.
\end{equation}

\subsection{Contrastive Learning Procedure}
The counterfactual mechanism is adopted to generate hard negative samples. After that, we need to conduct graph contrastive learning between the original graph and the perturbed graphs. In this paper, we follow a simple and widely-used graph contrastive learning schema to conduct it, which is \textit{dictionary look-up} method \cite{gcc}, shown in Figure \ref{fig:overview}. Given an original input graph $\mathcal{A}$, two negative graphs $\mathcal{A}'$ and $\tilde{\mathcal{A}}$, and two graph encoders, $g_p(\cdot)$ and $g_n(\cdot)$, we will have a sort of graph embeddings: $\mathbf{q}=g_p(\mathcal{A})$, $\mathbf{k}_+=\mathbf{k}_0=g_n(\mathcal{A})$, $\mathbf{k}_1=g_n(\mathcal{A}')$, and $\mathbf{k}_2=g_n(\tilde{\mathcal{A}})$. Specifically, the target graph will be encoded by both graph encoders, and $g_n(\cdot)$ will only be used to encode the generated hard negative samples. \textit{Dictionary look-up} method here tries to look up a single key (denoted by $\mathbf{k}_+$) that $\mathbf{q}$ matches in $\mathbb{K}$. Let $\mathbf{q}$ denotes the query key and $\mathbb{K}=\{\mathbf{k}_0, \mathbf{k}_1, \mathbf{k}_2\}$ be the dictionary, we take InfoNCE in \cite{infonce} to formulate contrastive learning procedure. So, we have the objective for the graph contrastive learning phase:
\begin{equation}
	\mathcal{L}_{contra}=-\log\frac{\exp(sim(\mathbf{q}, \mathbf{k}_+)/\tau)}{\sum^{|\mathcal{K}|-1}_{t=0}\exp(sim(\mathbf{q}, \mathbf{k}_t)/\tau)},
\end{equation}
where $\tau$ is the temperature hyperparameter.


After finishing the two training phases, including counterfactual mechanism-based hard negative samples generation and graph contrastive learning, we can obtain the trained embeddings of all nodes and graphs. The graph embeddings will be fed into a downstream prediction model to conduct graph classification tasks and evaluate the trained embeddings' quality.

\section{Experiment}
\begin{table*}[htbp]
\caption{Comparison experimental results. F1-micro and F1-macro scores (\%) with standard deviations are listed above.}
\label{tab:comparison}
\footnotesize
 \vspace{-3mm}
\begin{tabular}{c|cc|cc|cccc}
	\toprule
	{\color[HTML]{333333} }                                 & \multicolumn{2}{c|}{{\color[HTML]{333333} PROTEINS\_full}}                                        & \multicolumn{2}{c|}{{\color[HTML]{333333} FRANKENSTEIN}}                                          & \multicolumn{2}{c|}{{\color[HTML]{333333} Synthie}}                                                                 & \multicolumn{2}{c}{{\color[HTML]{333333} ENZYMES}}                                                \\ 
	\multirow{-2}{*}{{\diagbox{                                                                                                                
				Method}{Dataset}}} & {\color[HTML]{333333} F1-Micro}                 & {\color[HTML]{333333} F1-Macro}                 & {\color[HTML]{333333} F1-Micro}                 & {\color[HTML]{333333} F1-Macro}                 & {\color[HTML]{333333} F1-Micro}                 & \multicolumn{1}{c|}{{\color[HTML]{333333} F1-Macro}}              & {\color[HTML]{333333} F1-Micro}                 & {\color[HTML]{333333} F1-Macro}                 \\ \midrule
	{\color[HTML]{333333} RandomWalk}                       & {\color[HTML]{333333} -}                        & {\color[HTML]{333333} -}                        & {\color[HTML]{333333} 57.97(std 2.15)}          & {\color[HTML]{333333} 57.45(std 1.92)}          & {\color[HTML]{333333} 18.50(std 4.06)}          & \multicolumn{1}{c|}{{\color[HTML]{333333} 16.86(std 3.58)}}       & {\color[HTML]{333333} -}                        & {\color[HTML]{333333} -}                        \\
	{\color[HTML]{333333} ShortestPath}                     & {\color[HTML]{333333} 70.88(std 4.91)}          & {\color[HTML]{333333} 69.88(std 4.99)}          & {\color[HTML]{333333} 62.39(std 1.95)}          & {\color[HTML]{333333} 59.81(std 2.02)}          & {\color[HTML]{333333} 50.75(std 9.69)}          & \multicolumn{1}{c|}{{\color[HTML]{333333} 47.32(std 9.86)}}       & {\color[HTML]{333333} 27.83(std 6.37)}          & {\color[HTML]{333333} 27.18(std 5.93)}          \\
	{\color[HTML]{333333} GL}                               & {\color[HTML]{333333} 69.89(std 3.25)}          & {\color[HTML]{333333} 68.57(std 3.43)}          & {\color[HTML]{333333} 61.26(std 2.85)}          & {\color[HTML]{333333} 53.94(std 2.46)}          & {\color[HTML]{333333} 52.50(std 10.49)}         & \multicolumn{1}{c|}{{\color[HTML]{333333} 50.24(std 10.37)}}      & {\color[HTML]{333333} 31.67(std 7.03)}          & {\color[HTML]{333333} 30.02(std 7.13)}          \\
	{\color[HTML]{333333} WL}                               & {\color[HTML]{333333} 72.32(std 3.11)}          & {\color[HTML]{333333} 71.36(std 3.41)}          & {\color[HTML]{333333} -}                        & {\color[HTML]{333333} -}                        & {\color[HTML]{333333} -}                        & \multicolumn{1}{c|}{{\color[HTML]{333333} -}}                     & {\color[HTML]{333333} 37.83(std 4.95)}          & {\color[HTML]{333333} 36.42(std 5.78)}          \\ \midrule
	{\color[HTML]{333333} sub2vec}                          & {\color[HTML]{333333} 70.17(std 2.06)}          & {\color[HTML]{333333} 66.26(std 0.44)}           & {\color[HTML]{333333} 54.97(std 1.80)}          & {\color[HTML]{333333} 46.83(std 4.00)}          & {\color[HTML]{333333} 29.75(std 4.67)}          & \multicolumn{1}{c|}{{\color[HTML]{333333} 22.07(std 3.75)}}       & {\color[HTML]{333333} 19.67(std 3.64)}          & {\color[HTML]{333333} 13.34(std 4.33)}          \\
	{\color[HTML]{333333} graph2vec}                        & {\color[HTML]{333333} 68.65(std 3.45)}          & {\color[HTML]{333333} 64.16(std 5.00)}          & {\color[HTML]{333333} 61.70(std 3.04)}          & {\color[HTML]{333333} 59.68(std 0.22)}           & {\color[HTML]{333333} 54.25(std 0.62)}          & \multicolumn{1}{c|}{{\color[HTML]{333333} 35.17(std 0.26)}}       & {\color[HTML]{333333} 25.67(std 4.84)}          & {\color[HTML]{333333} 22.41(std 5.04)}          \\ \midrule
	{\color[HTML]{333333} InfoGraph}                        & {\color[HTML]{333333} 71.61(std 4.67)}          & {\color[HTML]{333333} {\underline{70.48(std 5.06)}}}    & {\color[HTML]{333333} {\underline{63.57(std 2.12)}}}    & {\color[HTML]{333333} {\underline{62.95(std 2.20)}}}    & {\color[HTML]{333333} 54.5(std 8.05)}           & \multicolumn{1}{c|}{{\color[HTML]{333333} 54.17(std 7.87)}}       & {\color[HTML]{333333} 38.33(std 7.03)}          & {\color[HTML]{333333} 37.07(std 6.89)}          \\
	{\color[HTML]{333333} MVGRL}                            & {\color[HTML]{333333} 72.06(std 3.29)}          & {\color[HTML]{333333} 69.53(std 3.61)}          & {\color[HTML]{333333} 61.89(std 1.40)}          & {\color[HTML]{333333} 59.65(std 1.50)}          & {\color[HTML]{333333} {\underline{62.00(std 9.07)}}}    & \multicolumn{1}{c|}{{\color[HTML]{333333} {\underline{61.59(std 9.52)}}}} & {\color[HTML]{333333} {\underline{40.50(std 7.85)}}}    & {\color[HTML]{333333} {\underline{38.7(std 9.12)}}}     \\
	{\color[HTML]{333333} GraphCL}                          & {\color[HTML]{333333} {\underline{73.05(std 3.29)}}}    & {\color[HTML]{333333} \textbf{71.04(std 3.35)}} & {\color[HTML]{333333} 62.62(std 2.49)}          & {\color[HTML]{333333} 61.89(std 2.57)}          & {\color[HTML]{333333} 57.50(std 9.08)}          & \multicolumn{1}{c|}{{\color[HTML]{333333} 55.87(std 8.87)}}       & {\color[HTML]{333333} 33.67(std 4.58)}          & {\color[HTML]{333333} 33.46(std 4.96)}          \\
	{\color[HTML]{333333} GCA}                              & {\color[HTML]{333333} 71.71(std 4.40)}          & {\color[HTML]{333333} 69.59(std 4.44)}          & {\color[HTML]{333333} 63.20(std 1.70)}          & {\color[HTML]{333333} 62.17(std 1.57)}          & {\color[HTML]{333333} 52.25(std 5.18)}          & \multicolumn{1}{c|}{{\color[HTML]{333333} 43.27(std 9.85)}}       & {\color[HTML]{333333} 34.00(std 5.01)}          & {\color[HTML]{333333} 33.62(std 5.01)}          \\ \midrule
	{\color[HTML]{333333} CGC}                    & {\color[HTML]{333333} \textbf{73.48(std 4.90)}} & {\color[HTML]{333333} 70.03(std 5.75)}          & {\color[HTML]{333333} \textbf{64.93(std 1.98)}} & {\color[HTML]{333333} \textbf{63.25(std 2.04)}} & {\color[HTML]{333333} \textbf{63.75(std 6.91)}} & {\color[HTML]{333333} \textbf{63.23(std 6.71)}}                   & {\color[HTML]{333333} \textbf{47.50(std 6.25)}} & {\color[HTML]{333333} \textbf{46.99(std 6.30)}} \\ \bottomrule
\end{tabular}
 \vspace{-3mm}
\end{table*}

We conduct comparison experiments to show the superiority of the proposed method. Supplementary experimental results are given to analyze the properties of the proposed method. This section discloses sufficient experimental settings and datasets for readers to reproduce the experiments.

\subsection{Datasets}
To fully demonstrate the performances of the proposed method compared to baselines, we choose several public and widely-used datasets from TUDataset \cite{tudataset}. 
All the datasets are available on the webpage\footnote{https://ls11-www.cs.tu-dortmund.de/staff/morris/graphkerneldatasets}. 
Recall that the feature masking operation necessary for our proposed method is a hard negative sample generation procedure. Hence, the graph datasets we use must contain high-quality node features. We select four datasets, which are PROTEINS\_full \cite{protein, enzymes}, FRANKENSTEIN \cite{frank}, Synthie \cite{synthie}, and ENZYMES \cite{protein, enzymes}. The detailed statistics of four datasets are shown in Table \ref{tab:statistics}.

\subsection{Baselines}

To verify the effectiveness and superiority of the proposed framework, we compare it with several unsupervised learning methods in three categories: graph kernels, graph embedding methods, and graph contrastive learning methods. For graph kernel methods, we choose four different kernels, including \textbf{RandomWalk Kernel} \cite{randomwalk}, \textbf{ShortestPath Kernel} \cite{shortestpath}, \textbf{Graphlet Kernel} \cite{graphlet}, and \textbf{Weisfeiler-Lehman Kernel} \cite{wl}.
For graph embedding methods, we select two methods, including \textbf{sub2vec} \cite{sub2vec} and \textbf{graph2vec} \cite{graph2vec}.
Since the proposed method in this paper belongs to graph contrastive learning, it is important to compare our method to current state-of-the-art graph contrastive learning methods. We choose four impactful methods in the literature:
\begin{itemize}
    \item \textbf{InfoGraph} \cite{infograph} maximises the mutual information between the graph-level representation and the representations of substructures of different scales (e.g., nodes, edges, triangles) to acquire comprehensive graph embeddings.
    \item \textbf{MVGRL} \cite{mvgrl} utilises graph diffusion techniques to generate multiple views to form contrasting pairs.
    \item \textbf{GraphCL} \cite{graph-aug} adopts graph augmentations to obtain contrasting pairs and studies how to utilise these augmentations.
    \item \textbf{GCA} \cite{graph-ada-aug} is an updated version of GraphCL, which adaptively augment the graph with the centrality of nodes or edges instead of augmenting uniformly.
\end{itemize}

\subsection{Experimental Settings}
\begin{figure*}[htbp]
	\centering
	\subfigure[PROTEINS\_full]{
		\begin{minipage}[t]{0.24\linewidth}
			\centering
			\includegraphics[width=1\textwidth]{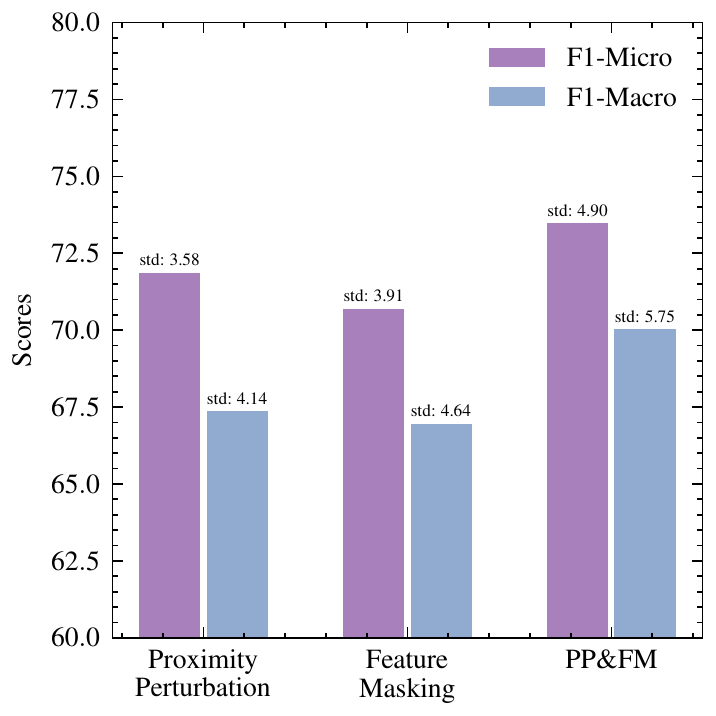}
		\end{minipage}%
	}%
	\subfigure[FRANKENSTEIN]{
		\begin{minipage}[t]{0.24\linewidth}
			\centering
			\includegraphics[width=1\textwidth]{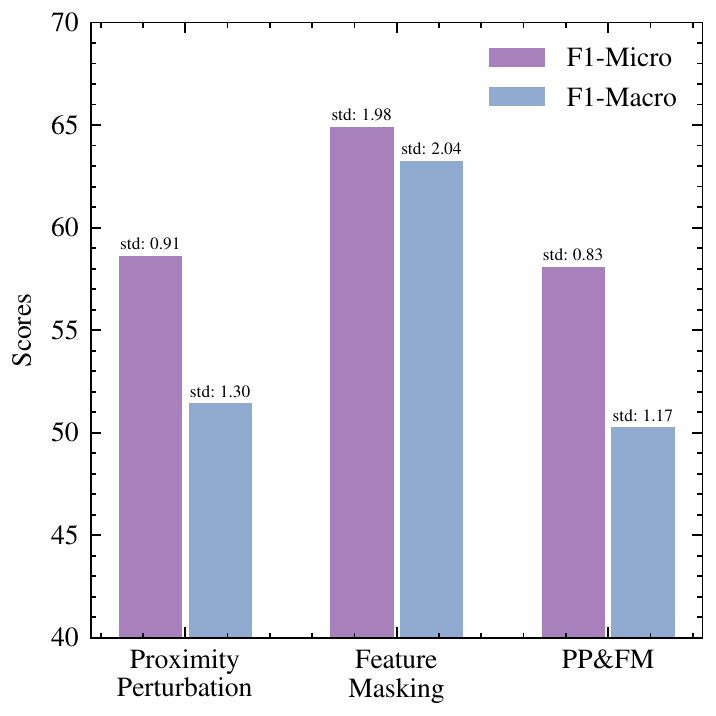}
		\end{minipage}%
	}%
	\subfigure[Synthie]{
		\begin{minipage}[t]{0.24\linewidth}
			\centering
			\includegraphics[width=1\textwidth]{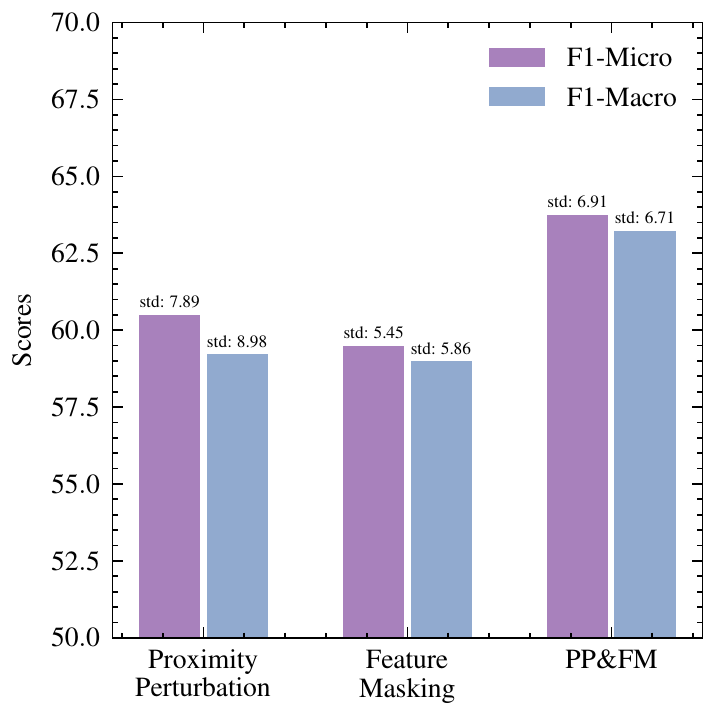}
		\end{minipage}
	}%
	\subfigure[ENZYMES]{
		\begin{minipage}[t]{0.24\linewidth}
			\centering
			\includegraphics[width=1\textwidth]{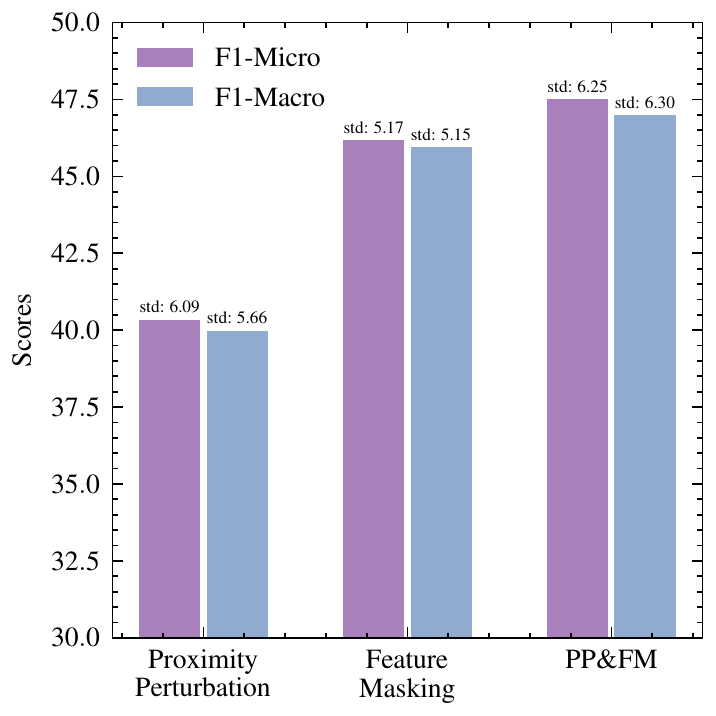}
		\end{minipage}
	}%
	\centering
	\caption{Graph classification results of the proposed method with different hard negative samples.}
        \Description{Graph classification results of the proposed method with different hard negative samples.}
	\label{fig:ablation-1}
 \vspace{-3mm}
\end{figure*}

For reproducibility, we introduce the detailed settings of the proposed method. For PROTEINS\_full, FRANKENSTEIN, and Synthie, we take GCN \cite{gcn} with three layers as the graph encoder. The learning rates for hard negative sample generation and contrastive learning are 0.0001. The training epochs for the two training stages are 80 and 30, respectively. For dataset ENZYMES, we adopt a 2-layer GIN \cite{gin} as the graph encoder. The learning rates for hard negative sample generation and contrastive learning are 0.001. The training epochs for both training stages are 100. The batch sizes for all the experiments are set to 256, while 128 is also feasible if GPU memory is limited for large graphs such as Synthie. The threshold $\omega$ and $\gamma$ mentioned previously are both 0.3. As to the temperature hyperparameter for contrastive learning, it is set to 1 for all the experiments.
We evaluate the proposed method via graph classification under the linear evaluation protocol. Specifically, we closely follow the evaluation protocol in InfoGraph and report the mean 10-fold cross-validation F1-Micro and F1-macro scores with standard deviation output by a linear SVM. 

\subsection{Comparison Experiment}
\begin{figure}[htbp]
	\centering
	\subfigure[F1-Micro, PROTEINS\_full]{
		\begin{minipage}[t]{0.38\linewidth}
			\centering
			\includegraphics[width=1\textwidth]{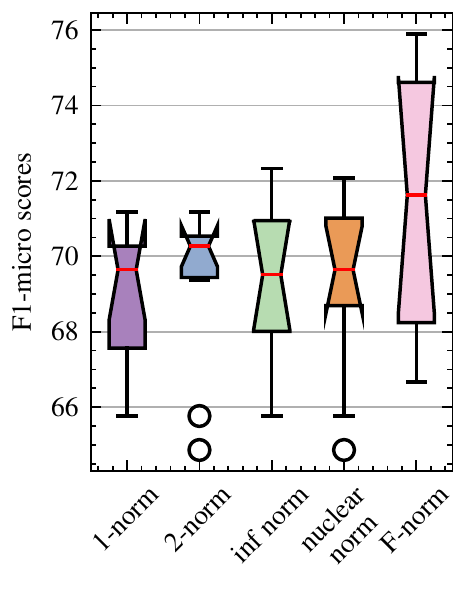}
		\end{minipage}%
	}%
	\subfigure[F1-Macro, PROTEINS\_full]{
		\begin{minipage}[t]{0.38\linewidth}
			\centering
			\includegraphics[width=1\textwidth]{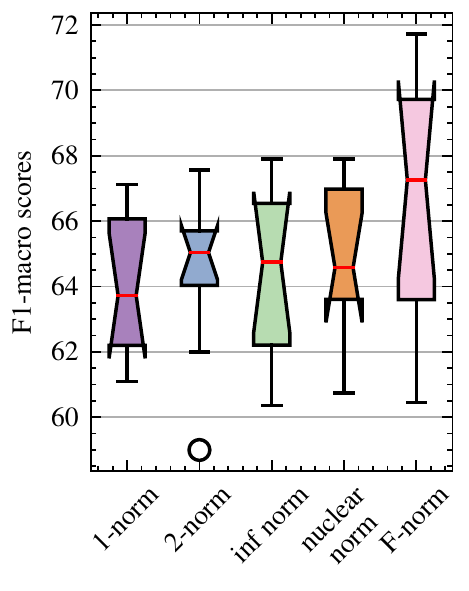}
		\end{minipage}%
	}%
	\\
	\subfigure[F1-Micro, FRANKENSTEIN]{
		\begin{minipage}[t]{0.38\linewidth}
			\centering
			\includegraphics[width=1\textwidth]{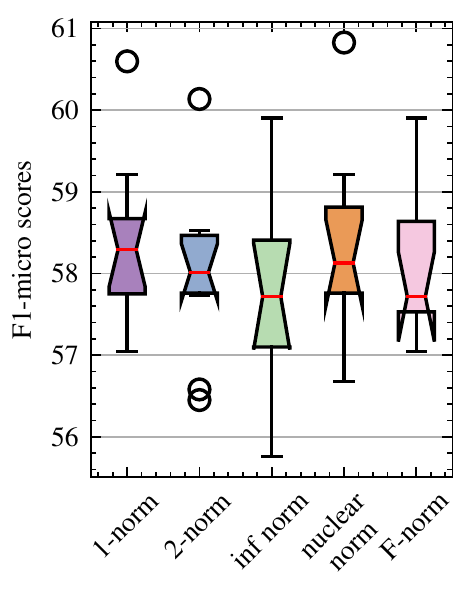}
		\end{minipage}%
	}%
	\subfigure[F1-Macro, FRANKENSTEIN]{
		\begin{minipage}[t]{0.38\linewidth}
			\centering
			\includegraphics[width=1\textwidth]{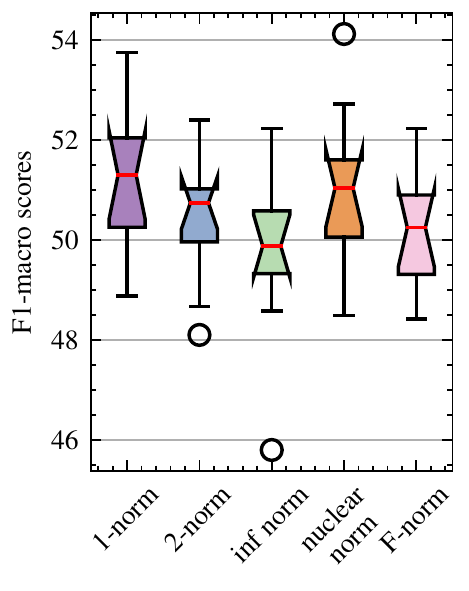}
		\end{minipage}%
	}%
	\\
	\subfigure[F1-Micro, Synthie]{
		\begin{minipage}[t]{0.38\linewidth}
			\centering
			\includegraphics[width=1\textwidth]{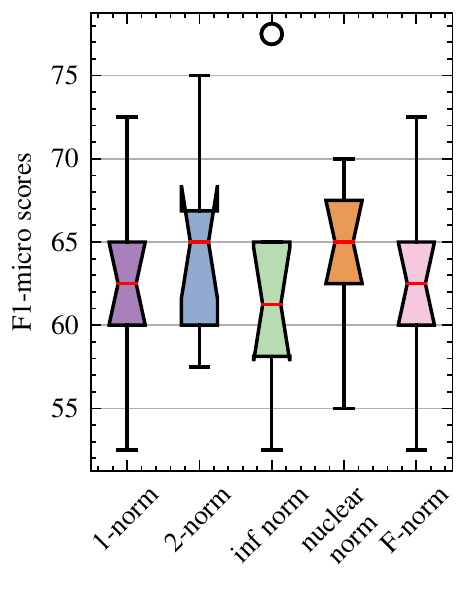}
		\end{minipage}%
	}%
	\subfigure[F1-Macro, Synthie]{
		\begin{minipage}[t]{0.38\linewidth}
			\centering
			\includegraphics[width=1\textwidth]{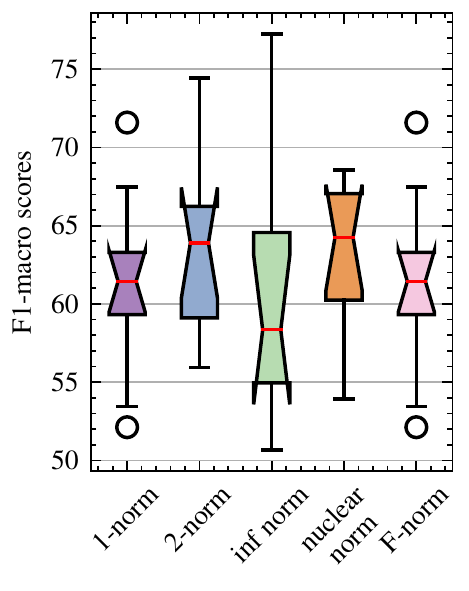}
		\end{minipage}%
	}%
	\\
	\subfigure[F1-Micro, ENZYMES]{
		\begin{minipage}[t]{0.38\linewidth}
			\centering
			\includegraphics[width=1\textwidth]{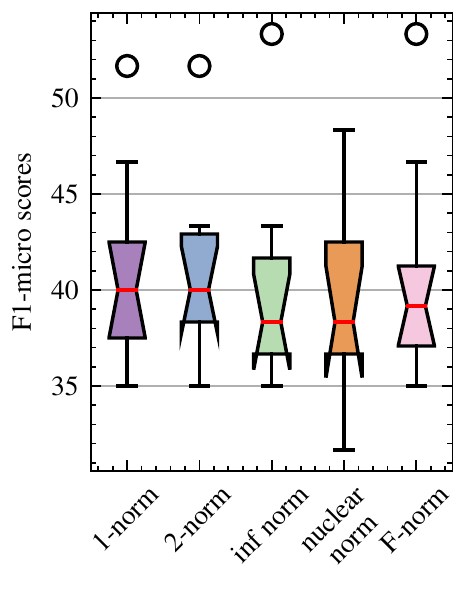}
		\end{minipage}%
	}%
	\subfigure[F1-Macro, ENZYMES]{
		\begin{minipage}[t]{0.38\linewidth}
			\centering
			\includegraphics[width=1\textwidth]{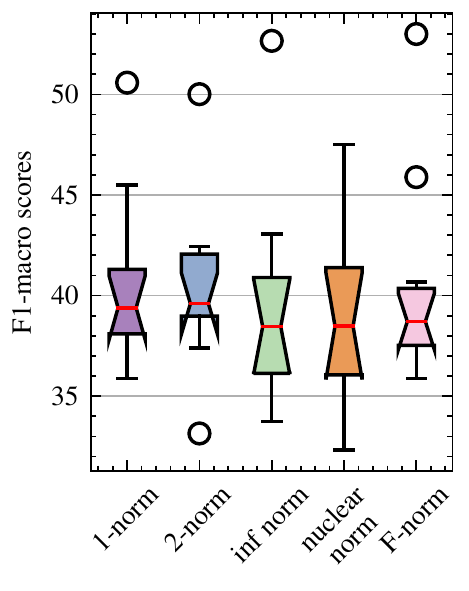}
		\end{minipage}%
	}%
	\centering
	\caption{The performances of the proposed method on all the datasets with hard negative samples, whose generation procedure adopts different matrix norms to measure similarity between the original input and the generated graphs.}
        \Description{The performances of the proposed method on all the datasets with hard negative samples, whose generation procedure adopts different matrix norms to measure similarity between the original input and the generated graphs.}
	\label{fig:ablation-2}
\end{figure}
The comparison experiment results for all baselines and our proposed method on all four datasets are shown in Table \ref{tab:comparison}. Generally, the proposed method outperforms the best baselines on all the datasets except PROTEINS\_full. Though our method has a mean F1-macro score lower than that of GraphCL, the gap between its F1-macro score and ours is insignificant as the standard deviation exists. We note that the proposed method significantly improves the dataset Synthie and ENZYMES. According to Table \ref{tab:statistics}, both of these datasets have multiple classes, which are 4 and 6, respectively. It indicates that the proposed method is superior in multiclass graph classification tasks. Recall one of the training objectives of hard negative samples generation, Equation (\ref{eq:hard}), which minimises the similarity between the probability distributions of which class the original graph and hard negative samples are. If there were a multiclass classification task, the Equation (\ref{eq:hard}) would minimise the similarity between two vectors (the vector refers to the probability distribution in our context) with more dimensions. Comparing two vectors with higher dimensionality could help the model to learn more information. So, it is reasonable that the proposed method has advantages on dataset Synthie and ENZYMES.

Graph kernel methods also achieve good performances compared to novel neural network methods. Nevertheless, some of them spend time on computing. Compared to our proposed method, they cannot be accelerated by GPUs, which is unaffordable under some real-world scenarios. Sub2Vec and Graph2Vec are two impactful graph embedding methods, which both leverage the idea of Word2Vec \cite{word2vec}. According to the experiment results, they cannot compete with the graph contrastive learning method, which is a novel and effective unsupervised graph learning paradigm nowadays.

Note that we selected four graph contrastive learning methods as baselines. All of them are impactful methods in the graph contrastive learning domain. InfoGraph is one of the first methods to introduce the idea of contrastive learning into the graph representation learning area. It achieved promising performances on several graph learning tasks. As shown in Table \ref{tab:comparison}, it has satisfying results on dataset PROTEINS\_full and FRANKENSTEIN. However, it may not be compatible with multiclass graph classification tasks, as the proposed CGC significantly outperforms it. Conversely, MVGRL performs better on Synthie and ENZYMES than on the other two datasets with only two classes. GCA is an updated version of GraphCL, and they share the same framework, but the improvement achieved by GCA is not significant. It has minor improvement on the dataset FRANKENSTEIN and ENZYMES. On dataset Synthie, it even has much worse performances. GraphCL and GCA try to conduct proper perturbations on the target to have positive or negative samples to form contrasting pairs. Specifically, GraphCL follows a random setting to perturb the graph, which cannot ensure the quality of the generated samples. GCA tries to adaptively locate the essential elements in the graph and perturb such identified elements according to their centrality. However, elements with high centrality are not always the critical factor determining the labels or semantics of the graph. Compared to our counterfactual hard negative samples generation method, these two methods have limitations in contrasting pairs generation.
We also note that GraphCL and GCA are incompatible with multiclass graph classification tasks. This is because the implementations of GraphCL and GCA both take the implementation of InfoGraph as the backbone. It is reasonable for them to have such a phenomenon.

\subsection{Ablation Study}
\subsubsection{\textbf{The impact on the graph contrastive learning with different types of generated hard negative samples.}}
Recall that we proposed to generate two different types of hard negative samples. In this section, we want to find out how much improvement can be brought by different hard negative samples. We conducted several experiments on the proposed method. We proceed with the graph contrastive learning procedure under three scenarios, which are 1) only to use the proximity perturbed graphs as a negative sample, 2) only to use the feature-masked graphs as negative samples, and 3) utilizing both types of graphs as negative samples. The results of the experiments are illustrated in Figure \ref{fig:ablation-1}. Utilizing both types of negative samples can achieve the best performances on all the datasets except FRANKENSTEIN. 
To achieve better results, utilising two hard negative samples can help the model capture the key semantics in proximity and feature space simultaneously. Moreover, we can form more contrasting pairs with more negative samples. Hence, the model can receive sufficient self-supervised signals to update parameters and perform better.

On dataset FRANKENSTEIN, the experiment results are not what we expected. The model trained only with the feature-masked graphs achieved the best performance. There is a significant gap between the performance of the model trained only with the proximity perturbed graphs and the model trained only with the feature-masked graphs. Such a gap makes the collaboration of two types of negative samples unsatisfying, resulting in the worse performance of the model trained with both types of generated negative samples.
Though the gap between the mean F1 scores of the model trained only with the proximity perturbed graphs and the model trained only with the feature-masked graphs on dataset ENZYMES is also significant, we note there is a larger standard deviation in the experimental results on the dataset ENZYMES. In this case, such a phenomenon indicates that the differences between the performances of the model trained only with the proximity perturbed graphs and the model trained only with the feature-masked graphs on dataset ENZYMES are not as significant as that on dataset FRANKENSTEIN.
According to Table \ref{tab:statistics}, graphs in dataset FRANKENSTEIN have much fewer nodes and edges than the other three datasets, but they have significantly larger node feature dimensionality. Masking features can bring more advantages to the model on dataset FRANKENSTEIN since the feature matrices are more complicated than the adjacency matrices. Such imbalance results in a considerable gap between the performances of the model trained only with the proximity perturbed graphs and the model trained only with the feature-masked graphs on dataset FRANKENSTEIN. We claim that perturbation to the aspects containing more informative semantics would bring more advantages to graph contrastive learning. Similar phenomena appear in the rest datasets. For example, graphs in dataset PROTEINS\_full and Synthie have complicated adjacency matrices with simple feature matrices. On these two datasets, the model trained only with the feature-masked graphs outperforms the model trained only with the proximity perturbed graphs.

\subsubsection{\textbf{How to measure the similarity in hard negative samples generation procedure?}}

\begin{table}[]
\caption{Five types of matrix norms and their definitions, where $M^*$ denotes the conjugate transpose of matrix $M\in\mathbb{R}^{n\times m}$, $\lambda_{max}(\cdot)$ is the function to have the largest eigenvalue of some matrix, and $tr(\cdot)$ represents trace function.}
	\label{tab:norms}
	\begin{tabular}{c|l|c}
		\toprule
		Matrix Norm  & \makecell[c]{Definition} & Complexity \\ \midrule
		1-norm       & $||M||_1=\max_{1\leq j\leq n}\sum^m_{i=1}|m_{ij}|$     & $\mathcal{O}(mn)$     \\ \midrule
		2-norm       & $||M||_2=\sqrt{\lambda_{max}(M^*M)}$     & $\mathcal{O}(m^3)$      \\ \midrule
		inf norm     & $||M||_{\infty}=\max_{1\leq i\leq m}\sum^n_{j=1}|m_{ij}|$     & $\mathcal{O}(mn)$      \\ \midrule
		nuclear norm & $||M||_*=tr(\sqrt{M^TM})$     & $\mathcal{O}(mn^2)$      \\ \midrule
		F-norm       & $||M||_F=\sqrt{\sum_{i=1}^{m}\sum_{j=1}^{n}|m_{ij}|^2}$     & $\mathcal{O}(mn)$      \\ \bottomrule
	\end{tabular}
  \vspace{-3mm}
\end{table}

Ensuring the generated negative samples have similar forms to the original input in proximity and feature space is the key to making the negative samples be \textit{hard}. A proper similarity measurement is important to achieve such a goal. In the methodology section, we introduced that we measure the similarity between the original input and the generated negative samples via calculating the norms of difference matrices $||A-A'_a||$ and $M'_b$. However, there are many different matrix norms. In this section, we examine the performances of the model trained with the negative samples in which different matrix norms were applied. We consider five different matrix norms shown in Table \ref{tab:norms}, and the experimental results are illustrated in Figure \ref{fig:ablation-2}.

Following the proposed generation protocol, the cost of \textit{2-norm} and \textit{nuclear norm} is much larger than others due to their computation complexity. Therefore, using these two matrix norms is not practical. As to \textit{inf norm}, its performances vary among all the datasets. The performances are not stable. Hence, \textit{1-norm} and \textit{F-norm} are more suitable for the similarity measurement in our hard negative samples generation protocol. Considering the simplicity of \textit{1-norm}, in most cases, it is a better option since it can achieve equivalent performances compared to \textit{F-norm}.

\section{Conclusion}
In this paper, we proposed a novel method, named CGC, to generate hard negative samples for graph contrastive learning. Compared to current graph contrastive learning methods and some classical graph kernel and graph embedding methods, it achieved state-of-the-art performances in most cases. We studied the effectiveness of the model trained with different types of generated hard negative samples. We found that perturbation made on the more complicated part of the graph data (e.g., node features or proximity) would bring more advantages to graph contrastive learning. Furthermore, we explore how to choose similarity measurement for hard negative sample generation from a perspective of matrix norm. There will be more methods to conduct such a task, and it would be interesting future work to improve the proposed method in this paper.

\begin{acks}
This work is supported by the Australian Research Council (ARC) under grant No. DP220103717, DP200101374, LP170100891, and LE220100078, and NSF under grants III-1763325, III-1909323, III-2106758, and SaTC-1930941. This research is also partially supported by APRC - CityU New Research Initiatives (No.9610565, Start-up Grant for New Faculty of City University of Hong Kong), SIRG - CityU Strategic Interdisciplinary Research Grant (No.7020046, No.7020074), HKIDS Early Career Research Grant (No.9360163), Huawei Innovation Research Program and Ant Group (CCF-Ant Research Fund).
\end{acks}

\bibliographystyle{ACM-Reference-Format}
\balance
\bibliography{ref}
\end{document}